\documentclass[10pt,twocolumn,letterpaper]{article}

\usepackage{cvpr}              

\usepackage{graphicx}
\usepackage{amsmath}
\usepackage{amssymb}
\usepackage{booktabs}
\usepackage{amsmath, bm}
\usepackage{pifont}
\newcommand{\cmark}{\ding{51}}%
\newcommand{\xmark}{\ding{55}}%

\usepackage[pagebackref,breaklinks,colorlinks]{hyperref}
\usepackage[accsupp]{axessibility}  

\usepackage[capitalize]{cleveref}
\crefname{section}{Sec.}{Secs.}
\Crefname{section}{Section}{Sections}
\Crefname{table}{Table}{Tables}
\crefname{table}{Tab.}{Tabs.}

\def\cvprPaperID{2046} 
\def\confName{CVPR}
\def\confYear{2022}

\begin{document}

\title{Spatial-Temporal Parallel Transformer for Arm-Hand Dynamic Estimation}

\author{Shuying Liu\footnotemark[1], Wenbin Wu\footnotemark[1], Jiaxian Wu, Yue Lin\\
NetEase Games AI Lab, Guangzhou, China\\
{\tt\small \{liushuying, wuwenbin02, wujiaxian, gzlinyue\}@corp.netease.com}
}

\twocolumn[{%
\renewcommand\twocolumn[1][]{#1}%
\maketitle

\begin{center}
    \centering
    \captionsetup{type=figure}
    \includegraphics[height=4.5cm]{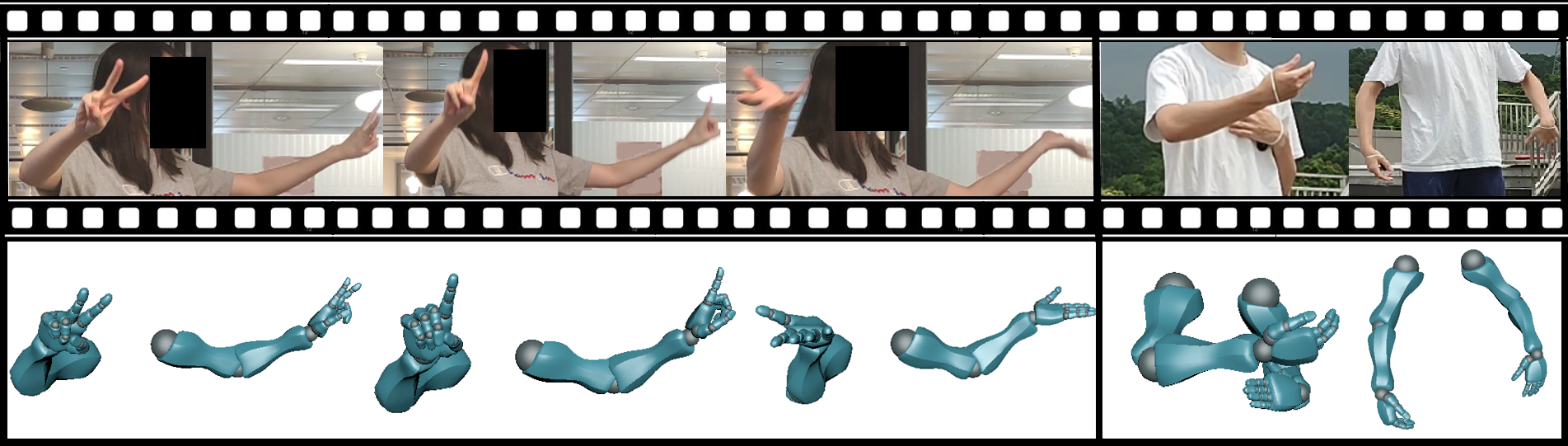}
    \captionof{figure}{We propose Spatial-Temporal Parallel Transformer to estimate arm and hand dynamics from monocular video by utilizing the arm-hand correlation as well as the temporal information. Row 1: Input video frames. Row 2: Estimated arm and hand dynamics of our method. The proposed method shows robustness under various challenging scenarios.}
\end{center}%
}]
\renewcommand{\thefootnote}{\fnsymbol{footnote}}
\footnotetext[1]{These authors contributed equally to this work.}
\begin{abstract}
      We propose an approach to estimate arm and hand dynamics from monocular video by utilizing the relationship between arm and hand. Although monocular full human motion capture technologies have made great progress in recent years, recovering accurate and plausible arm twists and hand gestures from in-the-wild videos still remains a challenge. To solve this problem, our solution is proposed based on the fact that arm poses and hand gestures are highly correlated in most real situations. To fully exploit arm-hand correlation as well as inter-frame information, we carefully design a Spatial-Temporal Parallel Arm-Hand Motion Transformer (PAHMT) to predict the arm and hand dynamics simultaneously. We also introduce new losses to encourage the estimations to be smooth and accurate. Besides, we collect a motion capture dataset including 200K frames of hand gestures and use this data to train our model. By integrating a 2D hand pose estimation model and a 3D human pose estimation model, the proposed method can produce plausible arm and hand dynamics from monocular video. Extensive evaluations demonstrate that the proposed method has advantages over previous state-of-the-art approaches and shows robustness under various challenging scenarios.

\end{abstract}

\section{Introduction}


Human arm-hand dynamics is an important part of full human motion capture, and can also be used for the control of human-machine interface.
However, although some methods are proposed to capture full human motion including hand gestures\cite{SMPL-X:2019,rong2020frankmocap,xiang2019monocular,zhang2021lightweight,zhou2021monocular}, most of them fail to take into account the correlation between arm and hand, treating body motion capture and gesture estimation as two separate tasks, resulting in inaccurate predictions in challenging scenarios. Zhou et al.\cite{zhou2021monocular} propose to learn human motion by considering body-hand correlation, but such correlation is only utilized in predicting 2D hand key-points and 3D body key-points, the final motion of body and hands are learned separately by different models. Ng et al.\cite{ng2021body2hands} introduce to learn the body-hand correlation to estimate conversational hand gestures. However, the method is restricted to the domain of conversational gesture prediction and cannot produce arm motions.



In this paper, we focus on the task of capturing accurate and plausible arm-hand dynamics from monocular video. 
Specifically, we introduce a spatial-temporal Parallel Arm-Hand Motion Transformer (PAHMT) to take full advantage of arm-hand correlations and inter-frame information. The estimation of 2D hand key-points and 3D arm key-points are first achieved by a light weight hand key-point detector and a 3D human key-point estimator respectively, which are used as input in our model. The PAHMT mainly consists of a spatial transformer and a temporal transformer. The spatial transformer is responsible for extracting the spatial feature, namely global correlations between arm and hand as well as local correlations between different joints, while the temporal transformer is designed to utilize inter-frame information. Besides, we introduce two losses to encourage the predictions to be smooth and accurate.
To train the proposed model, we collect a dataset of 200K frames of human motion (including hand gestures). Most of the collected sequences are dancing or sport motions that cover a lot of arm motions and hand gestures. We demonstrate that the proposed model can produce plausible estimation of arm-hand dynamics even in difficult scenarios such as occlusion or motion blur.

Our contributions can be summarized as follows:

\begin{itemize}
\item We propose to capture arm and hand dynamics simultaneously by leveraging the arm-hand correlations. By exploiting such correlations, the proposed model can make reasonable estimations of arm twists and hand gestures;
\item We design a spatial-temporal parallel transformer model to make full use of arm-hand correlation as well as inter-frame information, which enhances the robustness of the prediction; In addition, we introduce two losses to encourage smooth and accurate predictions;
\item Extensive evaluations demonstrate that the proposed method outperforms existing state-of-the-art approaches and shows robustness under various challenging scenarios.
\end{itemize}

\section{Related Work}
Great progress has been made on designing models to capture human pose or dynamic motions from visual observations\cite{gong2021poseaug,pavllo:videopose3d:2019,shan2021improving,li2021lifting,yan2018spatial,zhao2019semantic,cheng2019occlusion,cheng20203d,xiao2018simple}. Previous work always divide the task of full human motion capture into three independent sub-tasks, i.e., facial expression capture, body motion capture and hand gesture capture. Since our work focuses on capturing the dynamics of hands and arms (part of the body), we only present the related work as follows.


\subsection{Hand Pose Estimation}
\textbf{2D Hand Pose Estimation}. Some preceding work put much effort on building powerful datasets. Simon et al.\cite{simon2017hand} propose a hand key-point dataset with geometrically consistent annotations, i.e., annotations are provided even for occluded hand parts. With this dataset, a hand key-point detector is trained that can rival the performance of state-of-the-art hand key-point detectors using RGB-D inputs. To obtain more data for training, some researchers propose to first synthesize 3D hand data and then transfer them to real image style using GAN\cite{mueller2018ganerated,spurr2018cross}. Some work has been explored in network architecture design. Wang et al.\cite{wang2019srhandnet} introduce a novel architecture called SRHandNet for real-time 2D hand pose estimation from monocular RGB images, which can run at 40FPS while achieving competing results. Zimmermann et al.\cite{zimmermann2017learning} propose an architecture of four different network streams to make a full estimation of hand joints.

\textbf{3D Hand Pose Estimation.} The majority of state-of-the-art approaches predict 3D hand poses by recovering the hand meshes or regressing the coordinate of hands from single RGB image. Zhou et al.\cite{zhou2020monocular} use the estimated kinematic chain to estimate MANO\cite{romero2017embodied} coefficients and develop an inverse kinematics network to refine the predicted hand poses. Chen et al.\cite{chen2021camera} propose a camera-space mesh recovery framework to recover 3D hand mesh in camera-centered space. Kolotouros et al.\cite{kolotouros2019convolutional} directly regress the 3D coordinates of mesh vertices by training a graph convolutional network. On the other hand, some researchers have studied how to achieve hand pose estimation based on monocular video. Chen et al.\cite{chen2021temporal} exploit video temporal consistency to address the uncertainty caused by the lack of 3D joint annotations on training data. Apart from taking temporal sequences as input, Ng et al.\cite{ng2021body2hands} propose a novel learned deep prior of body motions for 3D hand shape synthesis and estimation in the domain of conversational gestures.

\subsection{Body Pose Estimation}
\textbf{Predicting 3D Joint Locations}. Most previous attempts on predicting 3D human joint locations take 2D human key-points as input and train models to learn the 2D to 3D lifting\cite{gong2021poseaug,pavllo:videopose3d:2019,li2021lifting,yan2018spatial,xiao2018simple}. Some of these work take 2D human pose of a single frame as input\cite{gong2021poseaug,xiao2018simple}, while more of them consider the inter-frame information and use a sequence of 2D key-points as input\cite{pavllo:videopose3d:2019,li2021lifting}. There are also some approaches that train models to directly estimate the 3D joint locations from visual input\cite{cheng2019occlusion,cheng20203d}. However, a common problem with these methods is that 3D skeletal rotations cannot be accurately recovered from 3D joint locations when the goal is to capture human skeletal motion, i.e., the twist component cannot be determined by joint locations.

\textbf{Predicting 3D Joint Rotations}. Most preceding work leverages parametric human body models for pose representation, e.g., SMPL\cite{loper2015smpl}, ADAM\cite{joo2018total}. A basic pipeline of these work\cite{kanazawa2018end} is to train a model to predict the parameters of SMPL model, such that the projected 2D joints on the input image can match the 2D joints detected by 2D pose estimators\cite{cao2017realtime,li2018crowdpose,xiu2018poseflow,fang2017rmpe}. Meanwhile, a discriminator is introduced to guide the model to produce realistic 3D poses. \cite{kocabas2019vibe,kanazawa2019learning} extend the baseline solution\cite{kanazawa2018end} for smoother results with video inputs. Nikos et al.\cite{kolotouros2019spin} propose to add an optimization process\cite{Bogo:ECCV:2016} after the CNN model. Nevertheless, without high quality $<image, motion>$ pair, these methods cannot directly utilize the motion capture data, i.e., motion capture data is always used to train the GAN part. Therefore, these methods show larger joint position error than methods that directly predict joint locations.

\subsection{Full Motion Capture}
There are also some work focusing on capturing full human motion\cite{joo2018total,xiang2019monocular,rong2020frankmocap,zhou2021monocular,SMPL-X:2019}. Rong et al. \cite{rong2020frankmocap} propose to capture full human motion by combining several existing technologies. Zhou et al.\cite{zhou2021monocular} utilize the body-hand correlation to train a network to predict 2D hand key-points and 3D body key-points, but key-points of hand and body are separately fed to different models for human motion. And SMPLify-X\cite{SMPL-X:2019} is an optimization process whose objective is to find a set of parameters of SMPL-X model that best fit the 2D key-points.

\section{Methodology}
\begin{figure*}[t]
\begin{center}
   \includegraphics[width=0.95\linewidth]{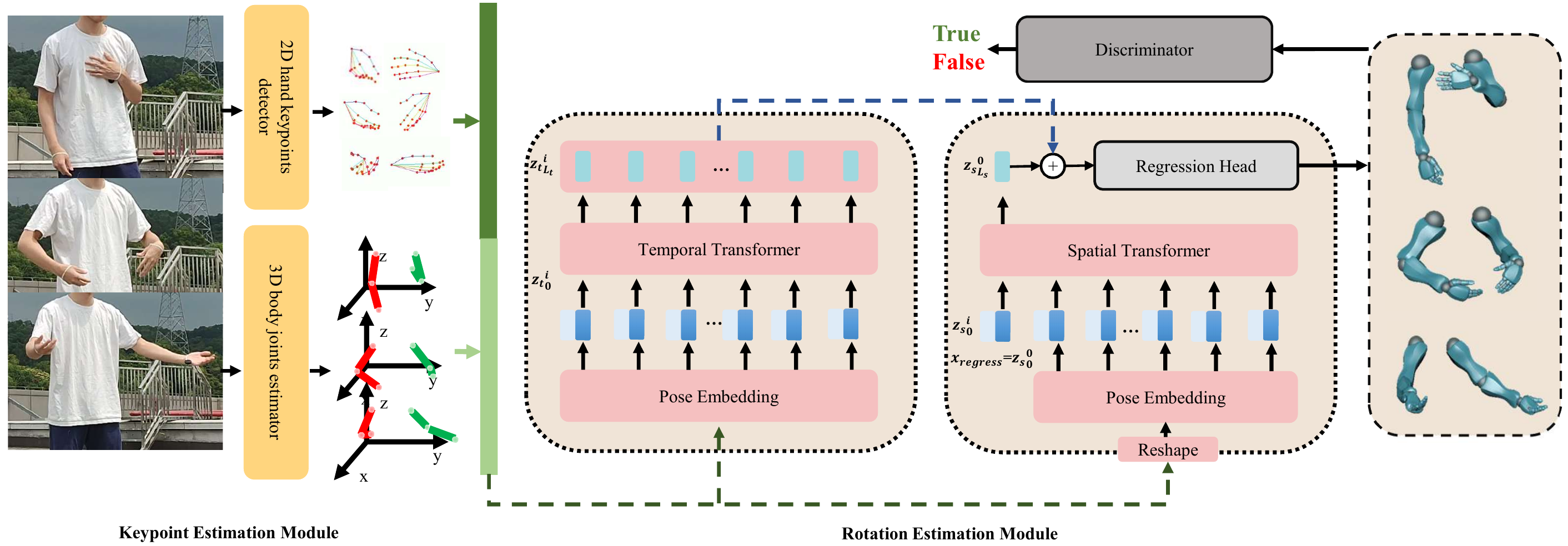}
\end{center}
   \caption{Overall pipeline of our method. The whole pipeline consists two modules, the key-point estimation module and the rotation estimation module. The key-point estimation module consists of a 2D hand key-point detector and a 3D body key-point estimator. The rotation estimation module consists of a carefully designed transformer-based network, which takes a sequence of 2D hand key-points and 3D arm joints as input, and gives a output prediction of arm-hand dynamics. Additionally, a discriminator is used to guide the network for plausible results.}
\label{fig:pipeline}
\end{figure*}

The goal of our work is to capture the dynamics of arms and hands from monocular video. It is important to note that some researchers use human joint locations as human poses, while others use skeletal rotation as human poses. In this work, we primarily focus on training models to estimate rotations of arms and hands. It is difficult to learn the image-to-rotation mapping directly. Meanwhile, learning such mapping requires motion capture data with corresponding video frames, which is hard to acquire. Therefore, we decouple this task into two sub-tasks, namely the key-point estimation task and the skeletal rotation estimation task. The overall pipeline of this work is illustrated in Fig \ref{fig:pipeline}. As for the key-point estimation module, two models are trained to predict 2D hand key-points and 3D body key-points respectively. Then, the rotation estimation module takes the predicted 2D hand key-points and 3D arm key-points (obtained from 3D body key-points) as inputs and trains a model to learn the 3D rotations of arms and hands. In the following we will briefly introduce the key-point estimation module while delving into the details of the rotation estimation module.

\subsection{Problem Definition}
The goal of this system $\mathcal{F}$ is to capture the rotations of arms and hands $\mathbf{Y}$ from visual observations $\mathbf {I}$. 

\begin{equation}
    \mathbf {Y} \bm{=} \mathcal{F}(\mathbf {I})
\end{equation}

The system is decoupled into two sub-modules, i.e., key-point estimation module $\mathcal{K}$ and rotation estimation module $\mathcal{R}$,

\begin{equation}
    \mathbf {K} \bm{=} \mathcal{K}(\mathbf {I})
\end{equation}

\begin{equation}
    \mathbf {Y} \bm{=} \mathcal{R}(\mathbf{K})
\end{equation}

where $\bm{\mathbf{Y}}=\{ \bm{y}_{0}, \bm{y}_{1}, ..., \bm{y}_{t},..., \bm{y}_{T}  \}$, $\bm{\mathbf{K}=} \{ \bm{k}_{0}, \bm{k}_{1}, ..., \bm{k}_{t},..., \bm{k}_{T}   \}$, $\bm{\mathbf{I}=} \{ \bm{i}_{0}, \bm{i}_{1}, ..., \bm{i}_{t},..., \bm{i}_{T}  \}$. $\bm{y}_{t} \in \bm{\mathbb{R}} ^{48\times 3}$ denotes the 3D rotation of arms and hands at time $t$, which is represented by 3D axis-angle. $\bm{k}_{t} \in \bm{\mathbb{R}} ^{4\times 3 + 42 \times 2}$ denotes the 3D local location of arm joints and 2D pixel coordinates of hands at time $t$. $\bm{i}_{t}$ denotes input image frame at time $t$.


\textbf{Difference with Full Body Inverse Kinematics}\cite{yamane2003natural,mistry2008inverse}: Full-body inverse kinematics (FBIK) is an optimization process which optimizes skeletal motions to fit 3D joints. Thus FBIK requires accurate 3D hand key-points and body key-points. Inaccurate 3D key-points might lead to implausible results. However, it’s much more difficult to obtain accurate 3D hand key-points than to obtain accurate 2D hand key-points from RGB image. Moreover, frame-by-frame IK process might cause large jitter in the results. In contrast, our solution does not require 3D hand key-points to produce smooth and plausible estimations from video input.


\subsection{Key-point Estimation Module}
\textbf{2D Hand Key-point Estimation}. To obtain accurate 2D hand key-points efficiently, we first crop the images of the hand regions according to detected 2D wrist joints, and then feed them to our network. The backbone of our model is based on the architecture of Mobilenetv3\cite{howard2019searching}. 
The model is trained to fit the ground truth probability map using L2 loss. Detailed configuration of this module can be found in the supplementary materials. 

\textbf{3D Body Key-point Estimation.} We use the VPose3D model\cite{pavllo:videopose3d:2019} trained on AMASS dataset\cite{mahmood2019amass} to obtain 3D body key-points, and use arm key-points as input of the rotation estimation module.

\subsection{Rotation Estimation Module}
To well utilize temporal information for better performance on capturing arm and hand dynamics, our model takes a sequence of key-points as input. Since transformer has been extensively proved to be efficient in various sequence-to-sequence tasks\cite{vaswani2017attention,lewis2019bart,song2019mass}, we design our network based on the transformer architecture.

\textbf{Arm-Hand Motion Transformer.} We employ a temporal transformer called Arm-Hand Motion Transformer (AHMT), as a baseline. The input pose sequence is treated the same way as tokens (words) in Natural Language Processing applications\cite{vaswani2017attention,lewis2019bart,song2019mass}. In order to project each token to a high dimensional space, a pose embedding module $\bm{\mathbf{E}}$ which consists of two convolution layers, is designed. And positional embedding $\bm{\mathbf{E_t}_{\text {pos }}}$ is used to retain positional information of the sequence. Given an input sequence, the input tokens of our transformer can be represented as follows:

\begin{align}
    \mathbf{z_t}_{0} &=\left[\mathbf{x}_{t}^{1} \mathbf{E} ; \mathbf{x}_{t}^{2} \mathbf{E} ; \cdots ; \mathbf{x}_{t}^{f} \mathbf{E}\right]+\mathbf{E_t}_{\text {pos }} \label{pos_emb}
\end{align}

where $X_t = [x_t^1;x_t^2;...;x_t^f]$ is the input sequence; $\mathbf{E} \in \mathbb{R}^{\left(J \cdot 3\right) \times D_t}$ is the pose embedding module; $\mathbf{E_t}_{\text {pos }} \in \mathbb{R}^{f \times D_t} $ is the positional embedding. $J$ is the number of total joints including arms and hands. $f$ is set to 32 during training;  $D_t=512$ is a constant latent vector size of our transformer.

Then a standard transformer encoder, consisting of alternating layers of Multi-head Self-Attention (MSA) and Multi-layer Perceptron (MLP) blocks (Eq.\ref{transformer_1},\ref{transformer_2)}), is employed to extract temporal information from these high dimensional features. We apply layer normalization (LN) before every block and residual connections after every block. The processing of our transformer encoder can be written as:

\begin{align}
    \mathbf{z_t}^{\prime}{ }_{\ell} &=\operatorname{MSA}\left(\operatorname{LN}\left(\mathbf{z_t}_{\ell-1}\right)\right)+\mathbf{z_t}_{\ell-1},  \ell=1 \ldots L_t \label{transformer_1}\\
    \mathbf{z_t}_{\ell} &=\operatorname{MLP}\left(\operatorname{LN}\left(\mathbf{z_t}^{\prime}{ }_{\ell}\right)\right)+\mathbf{z_t}_{\ell}^{\prime},  \ell=1 \ldots L_t \label{transformer_2)}
\end{align} 

  where $L_t$ represents the number of transformer encoder layer, and ${\mathbf{z_t}_{L_t}}$ is the last output of transformer encoder.

Finally, the arm-hand rotations are regressed by a regression head consisting of three convolution layers. We can get the output $\mathbf{y}$ as follows:

\begin{align}
\mathbf{y} &=\operatorname{Regression Head}\left(\mathbf{z_t}_{L_t}\right) \label{mlp}
\end{align}

 \textbf{Spatial-Temporal Parallel Arm-Hand Motion Transformer.} The AHMT baseline mainly focuses on extracting temporal features from input sequence. However, we believe that the spatial information, which represents the kinematic information between different joints and the correlations between arms and hands, is as important as the temporal information. Therefore, we design a Parallel Arm-Hand Motion Transformer (PAHMT) network to fuse the spatial and temporal information. As illustrated in Fig \ref{fig:pipeline}, the PAHMT mainly consists of two core components, the spatial transformer and the temporal transformer. The objective of the spatial transformer is to extract global correlations between arm poses and hand gestures, as well as the local correlations between different joints. Given an input sequence, we consider each frame, i.e., 3D direction vectors of arms and 2D coordinates of hands, as a token, and reshape the sequence $X_t \in \mathbb{R}^{f \times(J \cdot 3)}$ into $X_s \in \mathbb{R}^{f \times J \times 3}$ (Eq.\ref{reshape}). We conduct zero paddings on the 2D positions of hands, i.e., $(x, y)\to(x,y,0)$. Considering that both the global spatial correlations (e.g. waving arm always bring to waving hand) and the local spatial correlations (e.g. the movements of middle finger and ring finger are highly correlated) should be preserved across different frames, an extra learnable "regression token" ($\mathbf{z_s}_{0}^{0}$) is added to the sequence, whose state at the output of our spatial transformer encoder ($\mathbf{z_s}_{L_s}^{0}$) serves as the spatial representation, similar to the "classification token " proposed in ViT\cite{dosovitskiy2020image}. Each token is then fed into a convolution layer and added by a learnable position embedding. The resulting sequence of vectors is fed into a transformer encoder. 
The architecture of the temporal transformer is the same to AHMT (Eq.\ref{pos_emb},\ref{transformer_1},\ref{transformer_2)},\ref{mlp}). Formally, we have:

\begin{align}
    \mathbf{X_s} &=\operatorname{Reshape} \left(\operatorname{X_t}\right) \label{reshape}\\
    \mathbf{z_s}_{0} &=\left[\mathbf{x}_{\text {regress}} ; \mathbf{x}_{s}^{1} \mathbf{E} ; \mathbf{x}_{s}^{2} \mathbf{E} ; \cdots ; \mathbf{x}_{s}^{N} \mathbf{E}\right]+\mathbf{E_s}_{\text {pos }}\\
    \mathbf{z_s}^{\prime}{ }_{\ell} &=\operatorname{MSA}\left(\operatorname{LN}\left(\mathbf{z_s}_{\ell-1}\right)\right)+\mathbf{z_s}_{\ell-1}, \ell=1 \ldots L_s \\
    \mathbf{z_s}_{\ell} &=\operatorname{MLP}\left(\operatorname{LN}\left(\mathbf{z_s}^{\prime}{ }_{\ell}\right)\right)+\mathbf{z_s}_{\ell}^{\prime}, \ell=1 \ldots L_s 
\end{align}

where $N=48$ is the same as $J$, which represents the number of total joints including arms and hands. $\mathbf{E} \in \mathbb{R}^{3 \times D_s}$ is the pose embedding module of spatial transformer, $\mathbf{E_s}_{\text {pos }} \in \mathbb{R}^{N \times D_s}$ is the learnable positional encoding, $D_s=64$ is the constant latent vector size of our spatial transformer, $L_s$ is the number of the spatial transformer encoder layer.

Since the relationships between joints should be preserved across frames, only the output of the spatial transformer ($\mathbf{z_s}_{L_s}^{0}$) is taken as the spatial representation. Finally, we fuse the spatial features with the temporal features by element-wise adding and feed them to the regression head for the final predictions (Eq.\ref{reg_head}).
\begin{align}
    \mathbf{y} &=\operatorname{Regression Head}\left(\mathbf{z_t}_{L_t} + \mathbf{z_s}_{L_s}^{0}\right)\label{reg_head}
\end{align}

\subsection{Training}
In this section, we discuss the training details of the rotation estimation module.

\textbf{Pre-processing}: The 2D hand key-points are normalized. Specifically, the hand key-points are first subtracted by the coordinates of the wrist joints, then normalized by the bounding box of the corresponding hand. As for the frame in which hands cannot be detected, key-points of the previous frame are used. If the detector fails to detect hands in the first frame, we set the key-points of the first frame to be a zero vector. Considering that the predicted 3D skeletal joints might have different bone-length, for the 3D joints of arms, the direction vectors of upperarm and forearm are taken as input.

\textbf{Objective function}: As illustrated in Fig \ref{fig:pipeline}, the objective function of the rotation estimation module consists of two components. One is the reconstruction loss that guides the generator to fit the ground truth rotations of arms and hands. The other is the GAN loss\cite{goodfellow2014generative} introduced by the discriminator $\mathcal{D}$. Therefore, the full objective function can be formulated as:
\begin{equation}
     \underset{\mathcal{R}}{\mathrm{min} } ~\underset{\mathcal{D}}{\mathrm{max} } \bm{\mathfrak{L}}_{recon}(\mathcal{R}) + \lambda \bm{\mathfrak{L}}_{GAN}(\mathcal{R},\mathcal{D})
\end{equation}

where $\bm{\mathfrak{L}}_{GAN}$ is a basic form of GAN loss and $\lambda$ is the weight for this loss, $\bm{\mathfrak{L}}_{GAN}$ can be formulated as,

\begin{equation}
    \bm{\mathfrak{L}}_{GAN}(\mathcal{R},\mathcal{D}) \bm{=} \mathbb{E} [\mathrm{log} (\mathcal{D} (\mathbf{Y} ))] + \mathbb{E} [\mathrm{log} (1-\mathcal{R} (\mathbf{K} ))]
\end{equation}

and $\bm{\mathfrak{L}}_{recon}$ consists of three components. The basic one is a L1 loss, i.e.,

\begin{equation}
    \bm{\mathfrak{L}}_{L1}(\mathcal{R}) \bm{=} \left \| \mathcal{R}(\mathbf{K}) - \mathbf Y  \right \|_{1}
\end{equation}

One objective of this module is to produce smooth sequential results. Therefore a smooth loss is adopted, which can be regarded as a regularization term of the inter-frame difference, i.e.,
\begin{equation}
    \bm{\mathfrak{L}}_{smooth}(\mathcal{R}) \bm{=}  {\textstyle \sum_{t=1}^{T}} \left \| \mathcal{R}(\mathbf{\mathit{k_{t}}  }) - \mathcal{R}(\mathbf{\mathit{k_{t-1}}  })   \right \|_{1}
\end{equation}

It is noteworthy that rotations of different joints should have different importance in representing the poses. The rotation error of parent joints would propagate to all child joints, making the rotations of parent joints are more important than those of child joints.  Taking this into consideration, we introduce the FK loss. We first calculate the joint locations ${k_i^{*}}$ through the Forward Kinematic function $\mathcal{FK}$\cite{kucuk2006robot} of a pre-defined character skeleton, then compute the L2 loss on the ground truth joint location ${k_i^{*}}$ and the predicted joint location ${k_i}$,

\begin{equation}
    \bm{\mathfrak{L}}_{FK}(\mathcal{R}) \bm{=} \left \| \mathcal{FK}(\mathcal{R}(\mathbf{K})) - \mathbf K^{*}   \right \|_{2}
\end{equation}
$\mathbf K^{*}$ indicates 3D key-points of arms and hands, which can be obtained from motion capture data.

Finally, the full objective function $\bm{\mathfrak{L}}_{recon}$ for the generator can be defined as follows,

\begin{equation}
    \bm{\mathfrak{L}}_{recon} \bm{=} \bm{\mathfrak{L}}_{L1} + \gamma \bm{\mathfrak{L}}_{smooth} + \beta \bm{\mathfrak{L}}_{FK}
\end{equation}


\section{Experiments}
We analyse the performance of the proposed method via extensive experiments. We conduct ablation study to validate each individual component of our method and make both quantitative and perceptual comparisons with state-of-the-art approaches to demonstrate the advantage of our method.

\subsection{Dataset}

\textbf{Our Motion Capture Dataset:} Although there are some publicly available motion capture datasets\cite{mahmood2019amass,ionescu2013human3}, body motion data and hand motion data are always collected separately. We still lack a dataset of full human motion. We therefore collect a dataset of full human motion using motion capture devices. Specifically, we collect 500 sequences of full human motion data, with a total frame count of 200K. The dataset primarily consists of sequences of dancing or sports, covering a large number of body motions and hand gesture styles. We manually split the dataset into training set (90\%) and test set (10\%). We conduct ablation study on the test set. See the supplementary materials for samples of this dataset.


\textbf{In-the-wild 3D Body and Hand Gesture Dataset (BH Dataset):} To make comparison with the state-of-the-art gesture estimator\cite{ng2021body2hands}, we  conduct experiments on the dataset released by the author\cite{ng2021body2hands}. This dataset consists of hours of in-the-wild videos of 8 speakers and covers a wide range of gesture types. 

\textbf{Rendered Full Body Motion Dataset:} Although we already have full body motion capture data, we still lack in-the-wild data with full body motion annotations. Therefore, we propose a rendered dataset to simulate in-the-wild scenarios. Specifically, 10K frames of full body motion capture data are retargeting to 3 character models from MIXAMO\cite{Mixamo}. We compare our method with state-of-the-art full body motion capture methods\cite{rong2020frankmocap,choutas2020monocular} on this dataset. The samples of this dataset is shown in the supplementary materials.

\subsection{Implementation Details}
We first retarget all the skeletal motion sequences to the MIXAMO character model\cite{Mixamo}. Then we obtain the rotation (axis-angle) and the 3D position of each joint in world coordinate system. To generate training data, we project 3D locations of hands into the 2D camera plane to get the 2D pixel coordinates of hands. We apply a sliding window of size 32 with a step of 5 frames for each motion sequence, producing 30K training sequences. We train the network using Adam optimizer and we set the batch size,  weight decay and momentum to 128, 0.0005 and 0.95, respectively. The initial learning rate is set to 1e-3 and dropped by 50\% every 50 epochs. We train our model for 300 epochs. As for the objective functions, we empirically set $\lambda$, $\beta$, $\gamma$ to 0.05, 1.0, 1.0, respectively.
\subsection{Evaluation Metric}

As the goal of our work is to capture accurate skeletal rotations of arms and hands from visual input, besides the commonly used mean per joint position error (MPJPE, unit in meter), we also report the mean per joint rotation error (MPJRE), which measures the average absolute difference between predicted joint rotations and ground truth joint rotations. It is noteworthy that accurate hand rotations can be recovered from accurate 3D hand joints by inverse kinematic algorithms\cite{mistry2008inverse,yamane2003natural}, while arm rotations cannot be accurately recovered from 3D arm joints. Therefore, we use 1) the overall MPJPE; 2) MPJPE of hands; 3) MPJRE of arms to evaluate the performance.

\begin{table*}

\centering
\caption{Ablation study on different components of the proposed method. Evaluation is conducted on our motion capture dataset. We evaluate the overall MPJPE, MPJPE of hands and MPJRE of arms. Results indicate that each component shows positive impact on learning high quality arm and hand dynamics. h2h represents the models which take 2D hand key-points as input and predict hand gestures. ah2ah represents the models which take arm-hand key-points as input and predict arm-hand dynamics.}
\label{table:ablation}
\begin{tabular}{lccccccc}

\toprule
Architecture&h2h&ah2ah&Smooth Loss&FK Loss&MPJPE(hands)$\downarrow$ &MPJPE(overall)$\downarrow$ &MPJRE(arms)$\downarrow$ \\
\midrule
CNN&\cmark &\xmark&\xmark&\xmark&0.0300&-&-\\
CNN&\xmark&\cmark&\xmark&\xmark&0.0130&0.0517&0.0604\\
CNN&\xmark&\cmark&\cmark&\xmark&0.0133&0.0519&0.0603\\
CNN&\xmark&\cmark&\xmark&\cmark&0.0129&0.0413&0.0579\\
CNN&\xmark&\cmark&\cmark&\cmark&0.0127&0.0408&0.0577\\
AHMT&\xmark&\cmark&\cmark&\cmark&0.0098&0.0316&0.0445\\
PAHMT&\xmark&\cmark&\cmark&\cmark&\textbf{0.0087}&\textbf{0.0274}&\textbf{0.0375}\\
\bottomrule
\end{tabular}
\end{table*}



\begin{table}
\centering
\caption{Comparison with Body2Hands\cite{ng2021body2hands} on dataset released by \cite{ng2021body2hands}. We report MPJPE of hands for each method.}
\label{table:b2h}
\begin{tabular}{lc}
\toprule
Method&MPJPE(hands)$\downarrow$\\
\midrule
Body2Hands w/o image\cite{ng2021body2hands}&0.0422\\
Body2Hands w/ image\cite{ng2021body2hands}&0.0400\\
Body2Hands*\cite{ng2021body2hands}&0.0346\\
PAHMT(Ours)&\textbf{0.0281}\\
\bottomrule
\end{tabular}
\end{table}

\begin{table}
\centering
\caption{Comparison with state-of-the-art full body motion capture methods\cite{rong2020frankmocap,choutas2020monocular} on our rendered dataset. We report overall MPJPE and MPJRE of arms for each method.}
\label{table:render}
\begin{tabular}{lcc}
\toprule
Method&MPJPE(overall)$\downarrow$&MPJRE(arms)$\downarrow$\\
\midrule
FrankMocap\cite{rong2020frankmocap}&0.1556&0.2779\\
ExPose\cite{choutas2020monocular}&0.1530&0.2984\\
PAHMT(Ours)&\textbf{0.1375}&\textbf{0.1614}\\
\bottomrule
\end{tabular}
\end{table}

\subsection{Methods for comparison}
Firstly, we conduct ablation study to assess the contribution of each part of the method in estimating motions of arms and hands. Secondly, we compare our method with several state-of-the-art approaches\cite{ng2021body2hands,rong2020frankmocap,choutas2020monocular}. Note that our method simultaneously predicts the motion (rotations) of arms and hands, and the lack of a publicly available dataset with rotation annotations of both arms and hands prevents us from making a full comparison of both arms and hands rotations on an existing benchmark. Therefore, we only report the performance of hand pose estimation comparing our method with Body2Hands\cite{ng2021body2hands} on the dataset released by the authors. Furthermore, we make comparisons on our rendered dataset to evaluate the overall arm-hand dynamics with previous approaches. We also show results of our methods and state-of-the-arts approaches on some in-the-wild videos for perceptual evaluation. Specifically, we compare following methods:

\begin{itemize}
\item Body2Hands w/wo image\cite{ng2021body2hands}: State-of-the-art 3D hand shape synthesis and estimation by a learned deep prior of body motion. Models are re-trained using the codes released by the authors, since the pre-trained model is not available.

\item Body2Hands*\cite{ng2021body2hands}: The original Body2Hands model takes the output of MTC\cite{xiang2019monocular} as input and predicts the rotations of hands. However, in our experiments, MTC output is not provided. Thus we retrain a Body2Hands model of the same structure as the original version which takes 3D body joints and 2D hand key-points as input. We take the retrained version of Body2Hands as Body2Hands*.

\item ExPose\cite{choutas2020monocular}: State-of-the-art method for full body pose including hand gesture estimation, which directly regresses the body, face, and hand poses from RGB image. 

\item FrankMocap\cite{rong2020frankmocap}: State-of-the-art method for full body pose including hand gesture estimation, which leverages multiple leading solutions for pose estimation of different human body parts. 

\item Our CNN: The same model structure with Body2Hands\cite{ng2021body2hands}, trained with the proposed objective functions, takes 3D arm joints and 2D hand key-points as input and predicts the rotations of both arms and hands.

\item Our AHMT: Same configuration as Our CNN, except that the network architecture is replaced by a temporal transformer.

\item Our PAHMT: Same configuration as Our CNN, except that the network architecture is replaced by our carefully designed Spatial-Temporal Parallel Arm-Hand Motion Transformer(PAHMT). 

\end{itemize}
\begin{figure}[t]
\begin{center}
   \includegraphics[width=0.9\linewidth]{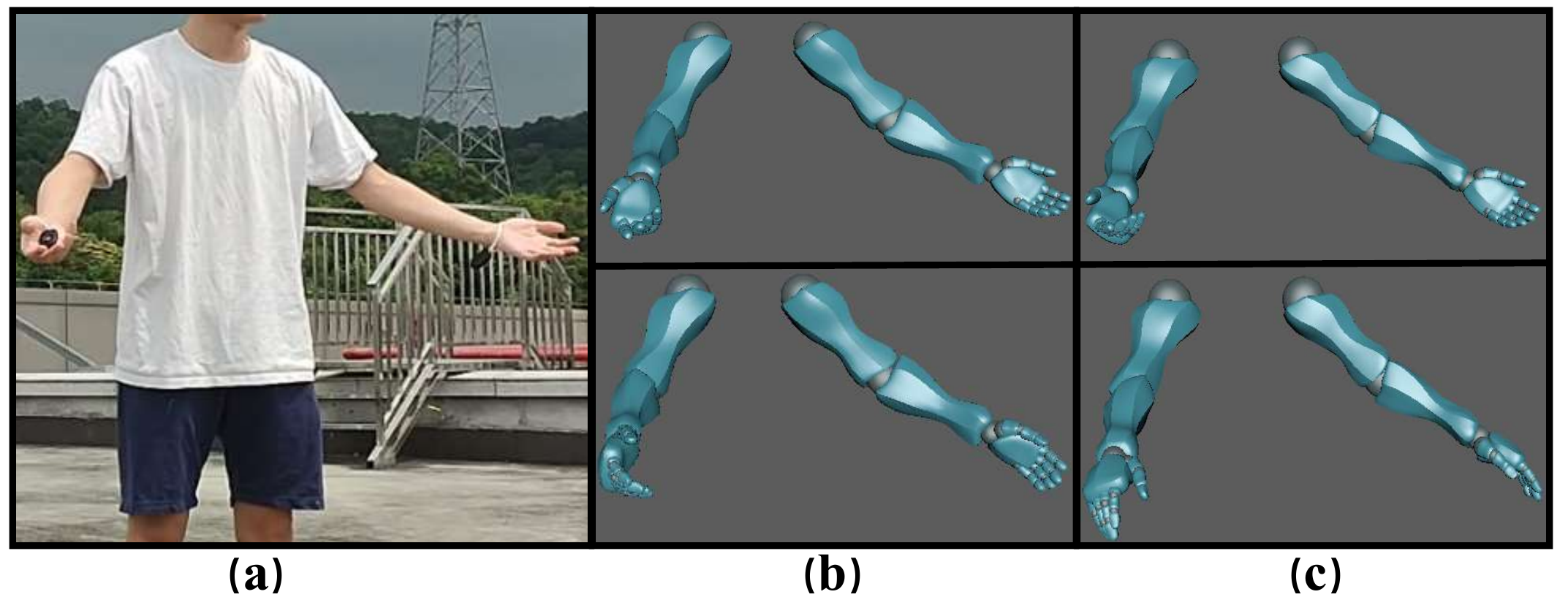}
\end{center}
   \caption{Effectiveness of leveraging arm-hand correlations. (a): Input frame. (b): Results of ah2ah. (c): Results of h2h+IK. To compare the rotations of arms, we set the rotations of wrists to zeros in the bottom row of (b) and (c).}
\label{fig:ahc}
\end{figure}
\begin{figure*}[t]
\begin{center}
   \includegraphics[height=8cm]{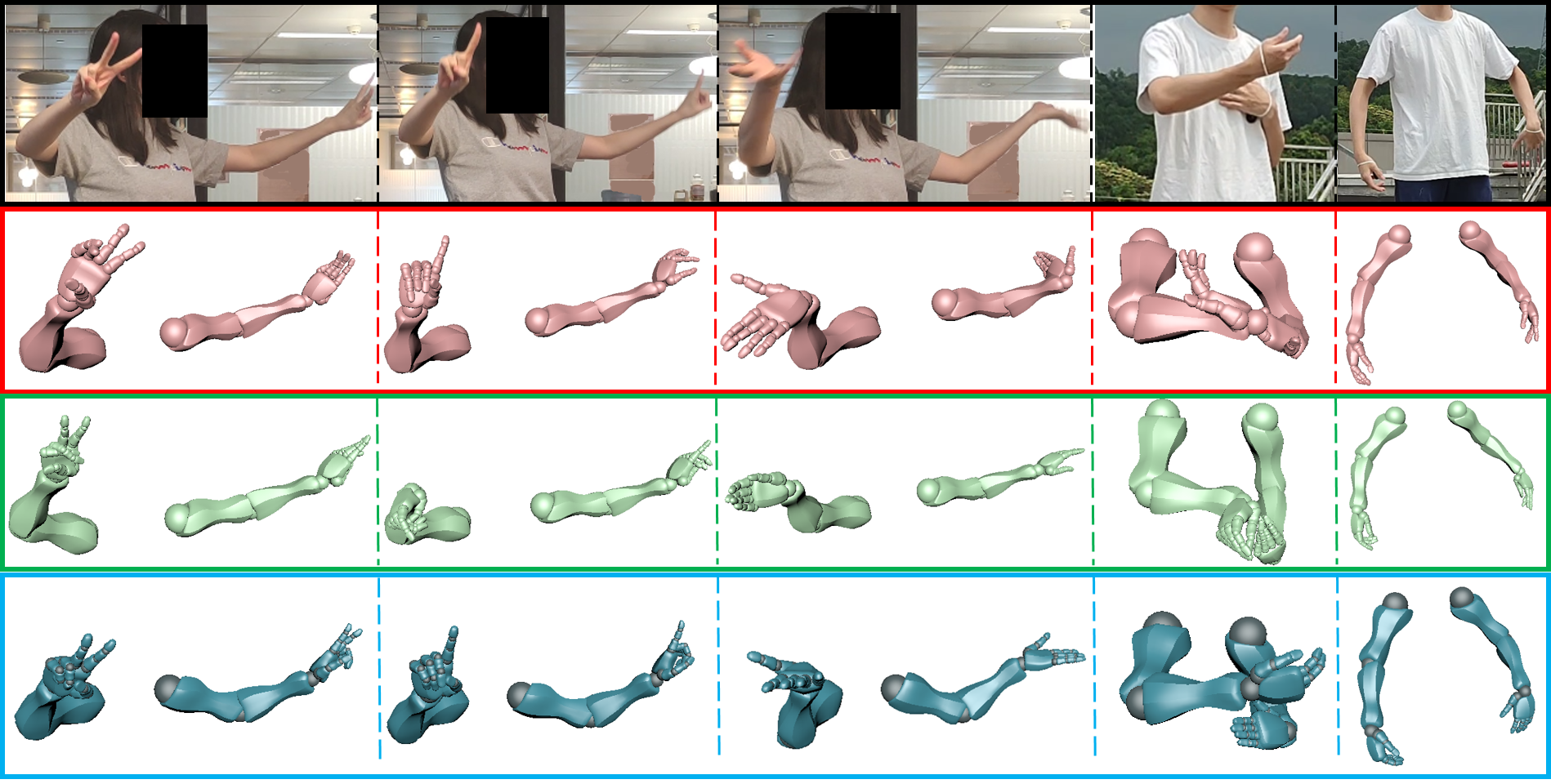}
\end{center}
   \caption{Visual comparison with state-of-the-art methods\cite{choutas2020monocular,rong2020frankmocap}. Row 1: In-the-wild video frames of various arm poses and hand gestures in challenging scenarios. Row 2: Results of FrankMocap\cite{rong2020frankmocap}. Row 3: Results of ExPose\cite{choutas2020monocular}. Row 4: Results of our method. Despite occlusion and motion blur, our method produces accurate and plausible results.}
\label{fig:compare}
\end{figure*}

\subsection{Ablation Study}
To evaluate the importance of each individual component of the proposed PAHMT and the impact of different hyperparameters of the network architecture, extensive experiments are conducted on our motion capture dataset.

\textbf{Baseline Setting}. We choose the network architecture of Our CNN as the baseline architecture. The objective function for the baseline model does not include $\bm{\mathfrak{L}}_{smooth}$ and $\bm{\mathfrak{L}}_{FK}$. For fair comparison, we fix the random seed for generating training batches in all the experiments. Adam optimizer is used  and we fix the batch size, learning rate, weight decay and momentum to 128, 1e-3, 0.0005 and 0.95 in all the experiments.

\textbf{Effectiveness of Leveraging Arm-hand Correlations}. As illustrated in Table \ref{table:ablation}, \textbf{h2h} (hand input to hand output) performs worse than \textbf{ah2ah} (arm-hand input to arm-hand output), indicating that the task of hand gesture estimation benefits from arm-hand correlation. Fig \ref{fig:ahc} intuitively shows the advantage of leveraging arm-hand correlations to predict arm-hand dynamics simultaneously. To demonstrate the superiority of \textbf{ah2ah} in predicting arm rotations, we compare \textbf{ah2ah} with \textbf{h2h+IK}\cite{yamane2003natural} (hand input to hand output, with IK to solve the arm rotations). We can observe that \textbf{ah2ah} gives more reasonable arm rotation predictions than \textbf{h2h+IK}.


\textbf{Effectiveness of the PAHMT Architecture}. Table \ref{table:ablation} presents the results. By making better use of inter-trame information, traditional temporal transformer (AHMT) significantly outperforms CNN (23\% error reduction on both MPJPE and MPJRE). We further investigate the impact
of introducing the spatial transformer. By utilizing the global correlation between arms and hands as well as the local correlation between different body joints, our PAHMT further improves the results of AHMT (13\% error reduction on MPJPE, 16\% error reduction on MPJRE). 

\textbf{Effectiveness of the Proposed Objective Functions}. We evaluate the impact of $\bm{\mathfrak{L}}_{smooth}$ and $\bm{\mathfrak{L}}_{FK}$. Table \ref{table:ablation} shows the results, $\bm{\mathfrak{L}}_{FK}$ shows advantage in reducing both the MPJPE and MPJRE. It is noteworthy that $\bm{\mathfrak{L}}_{smooth}$ is originally designed for temporal smoothness, not for the error reduction, therefore $\bm{\mathfrak{L}}_{smooth}$ shows no superiority in reducing the error when used independently. However, $\bm{\mathfrak{L}}_{smooth}$ also makes contribution to the accuracy improvement when integrated with $\bm{\mathfrak{L}}_{FK}$.

\textbf{Architecture Parameter Analysis}. We investigate different parameters of network architecture to search for a optimal setting. Detailed results can be found in the supplementary materials.

\subsection{Comparison with State-of-the-Arts}
Table \ref{table:b2h} shows the results on BH dataset. For fair comparison, the input 3D joints for our model is derived from the MTC\cite{xiang2019monocular} output. We can see that our PAHMT significantly outperforms the Body2Hands methods\cite{ng2021body2hands}. Note that BH dataset is annotated by MTC\cite{xiang2019monocular}, which means the label is not the ground truth. We show that our method can produce more plausible and smoother results than the annotations in supplementary materials.  

Table \ref{table:render} shows the results on the rendered dataset, which indicates that our method significantly outperforms state-of-the-art full body motion capture methods\cite{rong2020frankmocap,choutas2020monocular} on both the metric of MPJPE and MPJRE. It is worth noticing that our approach is significantly superior to existing methods in predicting arm rotations, which might indicate that our approach can make much better use of arm-hand correlation than state-of-the-arts.

\subsection{Visual Comparison}

To demonstrate the advantage of the proposed method, we perform experiments on in-the-wild videos for visual comparison with ExPose\cite{choutas2020monocular} and FrankMocap\cite{rong2020frankmocap}. For comparison, we conduct evaluation on in-the-wild videos including cases of clear and challenging unclear hands, as well as various arm poses.

Fig \ref{fig:compare} shows the visual comparison. We can observe that both ExPose\cite{choutas2020monocular} and FrankMocap\cite{rong2020frankmocap} fail to predict plausible hand gestures from unclear hand images. Meanwhile, these two state-of-the-arts cannot accurately estimate the rotations of arms. In contrast, our method produces the most accurate and plausible results even in challenging scenarios. More results can be found in the supplementary video.

\section{Discussion}
We propose a novel approach to estimate arm and hand dynamics from monocular video by utilizing the correlations between arm and hand motions. To make full use of the inter-frame information as well as the arm-hand correlations, a spatial-temporal parallel transformer model, PAHMT is proposed. To obtain smooth and accurate prediction, we propose new objective functions. Extensive experiments demonstrate the advantage of our method over previous state-of-the-arts. As for the limitation, our solution requires a motion capture dataset of both body motions and hand gestures to train the model. Future work may include leveraging the correlation of more body parts for high quality full body dynamic estimation and leveraging various dataset of annotations of different body parts. 

{\small
\bibliographystyle{ieee_fullname}
\bibliography{egbib}

\begin{thebibliography}{10}\itemsep=-1pt

\bibitem{Mixamo}
Mixamo. animate 3d characters for games, film, and more.
\newblock \url{http://https://www.mixamo.com}.
\newblock Accessed: 2021-09-30.

\bibitem{Bogo:ECCV:2016}
Federica Bogo, Angjoo Kanazawa, Christoph Lassner, Peter Gehler, Javier Romero,
  and Michael~J. Black.
\newblock Keep it {SMPL}: Automatic estimation of {3D} human pose and shape
  from a single image.
\newblock In {\em Computer Vision -- ECCV 2016}, Lecture Notes in Computer
  Science. Springer International Publishing, Oct. 2016.

\bibitem{cao2017realtime}
Zhe Cao, Tomas Simon, Shih-En Wei, and Yaser Sheikh.
\newblock Realtime multi-person 2d pose estimation using part affinity fields.
\newblock In {\em Proceedings of the IEEE conference on computer vision and
  pattern recognition}, pages 7291--7299, 2017.

\bibitem{chen2021temporal}
Liangjian Chen, Shih-Yao Lin, Yusheng Xie, Yen-Yu Lin, and Xiaohui Xie.
\newblock Temporal-aware self-supervised learning for 3d hand pose and mesh
  estimation in videos.
\newblock In {\em Proceedings of the IEEE/CVF Winter Conference on Applications
  of Computer Vision}, pages 1050--1059, 2021.

\bibitem{chen2021camera}
Xingyu Chen, Yufeng Liu, Chongyang Ma, Jianlong Chang, Huayan Wang, Tian Chen,
  Xiaoyan Guo, Pengfei Wan, and Wen Zheng.
\newblock Camera-space hand mesh recovery via semantic aggregation and adaptive
  2d-1d registration.
\newblock In {\em Proceedings of the IEEE/CVF Conference on Computer Vision and
  Pattern Recognition}, pages 13274--13283, 2021.

\bibitem{cheng20203d}
Yu Cheng, Bo Yang, Bo Wang, and Robby~T Tan.
\newblock 3d human pose estimation using spatio-temporal networks with explicit
  occlusion training.
\newblock In {\em Proceedings of the AAAI Conference on Artificial
  Intelligence}, volume~34, pages 10631--10638, 2020.

\bibitem{cheng2019occlusion}
Yu Cheng, Bo Yang, Bo Wang, Wending Yan, and Robby~T Tan.
\newblock Occlusion-aware networks for 3d human pose estimation in video.
\newblock In {\em Proceedings of the IEEE/CVF International Conference on
  Computer Vision}, pages 723--732, 2019.

\bibitem{choutas2020monocular}
Vasileios Choutas, Georgios Pavlakos, Timo Bolkart, Dimitrios Tzionas, and
  Michael~J Black.
\newblock Monocular expressive body regression through body-driven attention.
\newblock In {\em European Conference on Computer Vision}, pages 20--40.
  Springer, 2020.

\bibitem{dosovitskiy2020image}
Alexey Dosovitskiy, Lucas Beyer, Alexander Kolesnikov, Dirk Weissenborn,
  Xiaohua Zhai, Thomas Unterthiner, Mostafa Dehghani, Matthias Minderer, Georg
  Heigold, Sylvain Gelly, et~al.
\newblock An image is worth 16x16 words: Transformers for image recognition at
  scale.
\newblock {\em arXiv preprint arXiv:2010.11929}, 2020.

\bibitem{fang2017rmpe}
Hao-Shu Fang, Shuqin Xie, Yu-Wing Tai, and Cewu Lu.
\newblock {RMPE}: Regional multi-person pose estimation.
\newblock In {\em ICCV}, 2017.

\bibitem{gong2021poseaug}
Kehong Gong, Jianfeng Zhang, and Jiashi Feng.
\newblock Poseaug: A differentiable pose augmentation framework for 3d human
  pose estimation.
\newblock In {\em Proceedings of the IEEE/CVF Conference on Computer Vision and
  Pattern Recognition}, pages 8575--8584, 2021.

\bibitem{goodfellow2014generative}
Ian Goodfellow, Jean Pouget-Abadie, Mehdi Mirza, Bing Xu, David Warde-Farley,
  Sherjil Ozair, Aaron Courville, and Yoshua Bengio.
\newblock Generative adversarial nets.
\newblock {\em Advances in neural information processing systems}, 27, 2014.

\bibitem{howard2019searching}
Andrew Howard, Mark Sandler, Grace Chu, Liang-Chieh Chen, Bo Chen, Mingxing
  Tan, Weijun Wang, Yukun Zhu, Ruoming Pang, Vijay Vasudevan, et~al.
\newblock Searching for mobilenetv3.
\newblock In {\em Proceedings of the IEEE/CVF International Conference on
  Computer Vision}, pages 1314--1324, 2019.

\bibitem{ionescu2013human3}
Catalin Ionescu, Dragos Papava, Vlad Olaru, and Cristian Sminchisescu.
\newblock Human3. 6m: Large scale datasets and predictive methods for 3d human
  sensing in natural environments.
\newblock {\em IEEE transactions on pattern analysis and machine intelligence},
  36(7):1325--1339, 2013.

\bibitem{joo2018total}
Hanbyul Joo, Tomas Simon, and Yaser Sheikh.
\newblock Total capture: A 3d deformation model for tracking faces, hands, and
  bodies.
\newblock In {\em Proceedings of the IEEE conference on computer vision and
  pattern recognition}, pages 8320--8329, 2018.

\bibitem{kanazawa2018end}
Angjoo Kanazawa, Michael~J Black, David~W Jacobs, and Jitendra Malik.
\newblock End-to-end recovery of human shape and pose.
\newblock In {\em Proceedings of the IEEE conference on computer vision and
  pattern recognition}, pages 7122--7131, 2018.

\bibitem{kanazawa2019learning}
Angjoo Kanazawa, Jason~Y Zhang, Panna Felsen, and Jitendra Malik.
\newblock Learning 3d human dynamics from video.
\newblock In {\em Proceedings of the IEEE/CVF Conference on Computer Vision and
  Pattern Recognition}, pages 5614--5623, 2019.

\bibitem{kocabas2019vibe}
Muhammed Kocabas, Nikos Athanasiou, and Michael~J. Black.
\newblock Vibe: Video inference for human body pose and shape estimation.
\newblock In {\em The IEEE Conference on Computer Vision and Pattern
  Recognition (CVPR)}, June 2020.

\bibitem{kolotouros2019spin}
Nikos Kolotouros, Georgios Pavlakos, Michael~J Black, and Kostas Daniilidis.
\newblock Learning to reconstruct 3d human pose and shape via model-fitting in
  the loop.
\newblock In {\em ICCV}, 2019.

\bibitem{kolotouros2019convolutional}
Nikos Kolotouros, Georgios Pavlakos, and Kostas Daniilidis.
\newblock Convolutional mesh regression for single-image human shape
  reconstruction.
\newblock In {\em Proceedings of the IEEE/CVF Conference on Computer Vision and
  Pattern Recognition}, pages 4501--4510, 2019.

\bibitem{kucuk2006robot}
Serdar Kucuk and Zafer Bingul.
\newblock {\em Robot kinematics: Forward and inverse kinematics}.
\newblock INTECH Open Access Publisher, 2006.

\bibitem{lewis2019bart}
Mike Lewis, Yinhan Liu, Naman Goyal, Marjan Ghazvininejad, Abdelrahman Mohamed,
  Omer Levy, Ves Stoyanov, and Luke Zettlemoyer.
\newblock Bart: Denoising sequence-to-sequence pre-training for natural
  language generation, translation, and comprehension.
\newblock {\em arXiv preprint arXiv:1910.13461}, 2019.

\bibitem{li2018crowdpose}
Jiefeng Li, Can Wang, Hao Zhu, Yihuan Mao, Hao-Shu Fang, and Cewu Lu.
\newblock Crowdpose: Efficient crowded scenes pose estimation and a new
  benchmark.
\newblock {\em arXiv preprint arXiv:1812.00324}, 2018.

\bibitem{li2021lifting}
Wenhao Li, Hong Liu, Runwei Ding, Mengyuan Liu, and Pichao Wang.
\newblock Lifting transformer for 3d human pose estimation in video.
\newblock {\em arXiv preprint arXiv:2103.14304}, 2021.

\bibitem{loper2015smpl}
Matthew Loper, Naureen Mahmood, Javier Romero, Gerard Pons-Moll, and Michael~J
  Black.
\newblock Smpl: A skinned multi-person linear model.
\newblock {\em ACM transactions on graphics (TOG)}, 34(6):1--16, 2015.

\bibitem{mahmood2019amass}
Naureen Mahmood, Nima Ghorbani, Nikolaus~F Troje, Gerard Pons-Moll, and
  Michael~J Black.
\newblock Amass: Archive of motion capture as surface shapes.
\newblock In {\em Proceedings of the IEEE/CVF International Conference on
  Computer Vision}, pages 5442--5451, 2019.

\bibitem{mistry2008inverse}
Michael Mistry, Jun Nakanishi, Gordon Cheng, and Stefan Schaal.
\newblock Inverse kinematics with floating base and constraints for full body
  humanoid robot control.
\newblock In {\em Humanoids 2008-8th IEEE-RAS International Conference on
  Humanoid Robots}, pages 22--27. IEEE, 2008.

\bibitem{mueller2018ganerated}
Franziska Mueller, Florian Bernard, Oleksandr Sotnychenko, Dushyant Mehta,
  Srinath Sridhar, Dan Casas, and Christian Theobalt.
\newblock Ganerated hands for real-time 3d hand tracking from monocular rgb.
\newblock In {\em Proceedings of the IEEE Conference on Computer Vision and
  Pattern Recognition}, pages 49--59, 2018.

\bibitem{ng2021body2hands}
Evonne Ng, Shiry Ginosar, Trevor Darrell, and Hanbyul Joo.
\newblock Body2hands: Learning to infer 3d hands from conversational gesture
  body dynamics.
\newblock In {\em Proceedings of the IEEE/CVF Conference on Computer Vision and
  Pattern Recognition}, pages 11865--11874, 2021.

\bibitem{SMPL-X:2019}
Georgios Pavlakos, Vasileios Choutas, Nima Ghorbani, Timo Bolkart, Ahmed A.~A.
  Osman, Dimitrios Tzionas, and Michael~J. Black.
\newblock Expressive body capture: 3d hands, face, and body from a single
  image.
\newblock In {\em Proceedings IEEE Conf. on Computer Vision and Pattern
  Recognition (CVPR)}, 2019.

\bibitem{pavllo:videopose3d:2019}
Dario Pavllo, Christoph Feichtenhofer, David Grangier, and Michael Auli.
\newblock 3d human pose estimation in video with temporal convolutions and
  semi-supervised training.
\newblock In {\em Conference on Computer Vision and Pattern Recognition
  (CVPR)}, 2019.

\bibitem{romero2017embodied}
Javier Romero, Dimitrios Tzionas, and Michael~J Black.
\newblock Embodied hands: Modeling and capturing hands and bodies together.
\newblock {\em ACM Transactions on Graphics (ToG)}, 36(6):1--17, 2017.

\bibitem{rong2020frankmocap}
Yu Rong, Takaaki Shiratori, and Hanbyul Joo.
\newblock Frankmocap: Fast monocular 3d hand and body motion capture by
  regression and integration.
\newblock {\em arXiv preprint arXiv:2008.08324}, 2020.

\bibitem{shan2021improving}
Wenkang Shan, Haopeng Lu, Shanshe Wang, Xinfeng Zhang, and Wen Gao.
\newblock Improving robustness and accuracy via relative information encoding
  in 3d human pose estimation.
\newblock {\em arXiv preprint arXiv:2107.13994}, 2021.

\bibitem{simon2017hand}
Tomas Simon, Hanbyul Joo, Iain Matthews, and Yaser Sheikh.
\newblock Hand keypoint detection in single images using multiview
  bootstrapping.
\newblock In {\em Proceedings of the IEEE conference on Computer Vision and
  Pattern Recognition}, pages 1145--1153, 2017.

\bibitem{song2019mass}
Kaitao Song, Xu Tan, Tao Qin, Jianfeng Lu, and Tie-Yan Liu.
\newblock Mass: Masked sequence to sequence pre-training for language
  generation.
\newblock {\em arXiv preprint arXiv:1905.02450}, 2019.

\bibitem{spurr2018cross}
Adrian Spurr, Jie Song, Seonwook Park, and Otmar Hilliges.
\newblock Cross-modal deep variational hand pose estimation.
\newblock In {\em Proceedings of the IEEE Conference on Computer Vision and
  Pattern Recognition}, pages 89--98, 2018.

\bibitem{vaswani2017attention}
Ashish Vaswani, Noam Shazeer, Niki Parmar, Jakob Uszkoreit, Llion Jones,
  Aidan~N Gomez, {\L}ukasz Kaiser, and Illia Polosukhin.
\newblock Attention is all you need.
\newblock In {\em Advances in neural information processing systems}, pages
  5998--6008, 2017.

\bibitem{wang2019srhandnet}
Yangang Wang, Baowen Zhang, and Cong Peng.
\newblock Srhandnet: Real-time 2d hand pose estimation with simultaneous region
  localization.
\newblock {\em IEEE transactions on image processing}, 29:2977--2986, 2019.

\bibitem{xiang2019monocular}
Donglai Xiang, Hanbyul Joo, and Yaser Sheikh.
\newblock Monocular total capture: Posing face, body, and hands in the wild.
\newblock In {\em Proceedings of the IEEE/CVF Conference on Computer Vision and
  Pattern Recognition}, pages 10965--10974, 2019.

\bibitem{xiao2018simple}
Bin Xiao, Haiping Wu, and Yichen Wei.
\newblock Simple baselines for human pose estimation and tracking.
\newblock In {\em Proceedings of the European conference on computer vision
  (ECCV)}, pages 466--481, 2018.

\bibitem{xiu2018poseflow}
Yuliang Xiu, Jiefeng Li, Haoyu Wang, Yinghong Fang, and Cewu Lu.
\newblock {Pose Flow}: Efficient online pose tracking.
\newblock In {\em BMVC}, 2018.

\bibitem{yamane2003natural}
Katsu Yamane and Yoshihiko Nakamura.
\newblock Natural motion animation through constraining and deconstraining at
  will.
\newblock {\em IEEE Transactions on visualization and computer graphics},
  9(3):352--360, 2003.

\bibitem{yan2018spatial}
Sijie Yan, Yuanjun Xiong, and Dahua Lin.
\newblock Spatial temporal graph convolutional networks for skeleton-based
  action recognition.
\newblock In {\em Thirty-second AAAI conference on artificial intelligence},
  2018.

\bibitem{zhang2021lightweight}
Yuxiang Zhang, Zhe Li, Liang An, Mengcheng Li, Tao Yu, and Yebin Liu.
\newblock Lightweight multi-person total motion capture using sparse multi-view
  cameras.
\newblock In {\em Proceedings of the IEEE/CVF International Conference on
  Computer Vision}, pages 5560--5569, 2021.

\bibitem{zhao2019semantic}
Long Zhao, Xi Peng, Yu Tian, Mubbasir Kapadia, and Dimitris~N Metaxas.
\newblock Semantic graph convolutional networks for 3d human pose regression.
\newblock In {\em Proceedings of the IEEE/CVF Conference on Computer Vision and
  Pattern Recognition}, pages 3425--3435, 2019.

\bibitem{zhou2021monocular}
Yuxiao Zhou, Marc Habermann, Ikhsanul Habibie, Ayush Tewari, Christian
  Theobalt, and Feng Xu.
\newblock Monocular real-time full body capture with inter-part correlations.
\newblock In {\em Proceedings of the IEEE/CVF Conference on Computer Vision and
  Pattern Recognition}, pages 4811--4822, 2021.

\bibitem{zhou2020monocular}
Yuxiao Zhou, Marc Habermann, Weipeng Xu, Ikhsanul Habibie, Christian Theobalt,
  and Feng Xu.
\newblock Monocular real-time hand shape and motion capture using multi-modal
  data.
\newblock In {\em Proceedings of the IEEE/CVF Conference on Computer Vision and
  Pattern Recognition}, pages 5346--5355, 2020.

\bibitem{zimmermann2017learning}
Christian Zimmermann and Thomas Brox.
\newblock Learning to estimate 3d hand pose from single rgb images.
\newblock In {\em Proceedings of the IEEE international conference on computer
  vision}, pages 4903--4911, 2017.

\end{thebibliography}
}

\end{document}